\documentclass{article}

\usepackage{microtype}
\usepackage{amsfonts}
\usepackage{graphicx}
\usepackage{subfigure}
\usepackage{booktabs} 
\usepackage{algorithm}
\usepackage[noend]{algpseudocode}
\usepackage{hyperref}

\usepackage[accepted]{icml2019}

\icmltitlerunning{Collaborative Evolutionary Reinforcement Learning}

\begin{document}

\twocolumn[
\icmltitle{Collaborative Evolutionary Reinforcement Learning}



\icmlsetsymbol{equal}{*}

\begin{icmlauthorlist}
\icmlauthor{Shauharda Khadka}{Intel,OSU}
\icmlauthor{Somdeb Majumdar}{Intel}
\icmlauthor{Tarek Nassar}{Intel}
\icmlauthor{Zach Dwiel}{Intel}
\icmlauthor{Evren Tumer}{Intel}
\icmlauthor{Santiago Miret}{Intel}
\icmlauthor{Yinyin Liu}{Intel}
\icmlauthor{Kagan Tumer}{OSU}
\end{icmlauthorlist}

\icmlaffiliation{Intel}{Intel AI Lab}
\icmlaffiliation{OSU}{Collaborative Robotics and Intelligent Systems Institute, Oregon State University}

\icmlcorrespondingauthor{Shauharda Khadka}{shauharda.khadka@intel.com}
\icmlcorrespondingauthor{Somdeb Majumdar}{somdeb.majumdar@intel.com}

\icmlkeywords{Machine Learning, Deep Reinforcement Learning, Evolution, Meta-learning}

\vskip 0.3in
]



\printAffiliationsAndNotice{}  

\begin{abstract}

Deep reinforcement learning algorithms have been successfully applied to a range of challenging control tasks. However, these methods typically struggle with achieving effective exploration and are extremely sensitive to the choice of hyperparameters. One reason is that most approaches use a noisy version of their operating policy to explore - thereby limiting the range of exploration. In this paper, we introduce Collaborative Evolutionary Reinforcement Learning (CERL), a scalable framework that comprises a portfolio of policies that simultaneously explore and exploit diverse regions of the solution space. A collection of learners - typically proven algorithms like TD3 - optimize over varying time-horizons leading to this diverse portfolio. All learners contribute to and use a shared replay buffer to achieve greater sample efficiency. Computational resources are dynamically distributed to favor the best learners as a form of online algorithm selection. Neuroevolution binds this entire process to generate a single emergent learner that exceeds the capabilities of any individual learner. Experiments in a range of continuous control benchmarks demonstrate that the emergent learner significantly outperforms its composite learners while remaining overall more sample-efficient - notably solving the Mujoco Humanoid benchmark where all of its composite learners (TD3) fail entirely in isolation.

\end{abstract}

\section{Introduction}
\label{sec:intro}

Reinforcement learning (RL) has been successfully applied to a number of challenging tasks, ranging from arcade games \cite{mnih2015,mnih2016}, board games \cite{silver2016} to robotic control \cite{andrychowicz2017, lillicrap2015}. A driving force behind the explosion of RL applications is its integration with powerful non-linear function approximators like deep neural networks. This partnership, often referred to as Deep Reinforcement Learning (DRL), has enabled RL to successfully extend to tasks with high-dimensional input and action spaces. However, widespread adoption of these techniques to real-world problems is still limited by two major challenges: the difficulty in achieving effective exploration and brittle convergence properties that require careful tuning of the hyperparameters by a designer.

First, exploration is a key component for successful reinforcement learning. It enables an agent to learn good policies and avoid converging prematurely to local optima. Designing exploration strategies that lead to a diverse set of experiences remains a key challenge for DRL operating on high dimensional action and state spaces \cite{plappert2017}. Many methods have been formulated to address this issue, ranging from intrinsic motivation \cite{bellemare2016}, count-based exploration \cite{ostrovski2017, tang2017}, curiosity \cite{pathak2017} and variational information maximization \cite{houthooft2016}. Further, a parallel class of techniques emphasize exploration by adding noise directly to the parameters of the agent \cite{fortunato2017, plappert2017, khadka2018evolution}. However, each of these techniques either relies on supplementary structures or introduces task-specific parameters that need to be tuned rigorously. A general exploration strategy that is universally applicable across tasks and learning algorithms remains an active area of research. 

Second, DRL approaches are often notoriously sensitive to their hyperparamaters \cite{henderson2017,islam2017} and demonstrate brittle convergence properties \cite{haarnoja2018}. This is particularly true for off-policy approaches that use a replay buffer to leverage past experiences \cite{bhatnagar2009,duan2016}. In part, this sensitivity is coupled with the difficulty of effective exploration. A standard reinforcement learner employs a noisy version of its operating policy as its behavioral policy for exploration. This puts the burden of both exploitation and exploration onto the same set of hyperparameters. 

\begin{figure}[t]
    \centering
    \includegraphics[width=\linewidth,keepaspectratio]{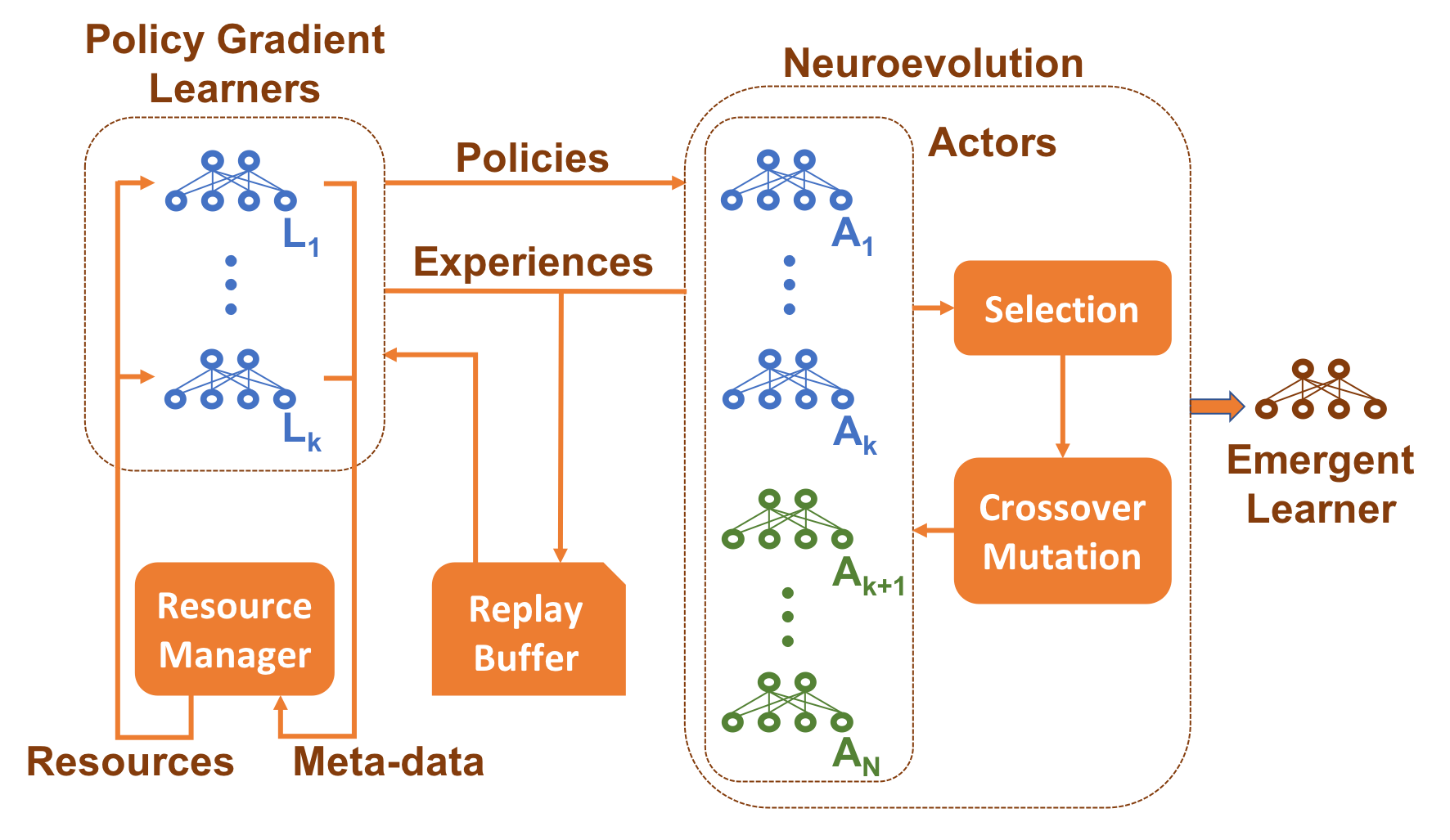}
    \caption{High level schematic of CERL. A portfolio of policy gradient learners operate in parallel to neuroevolution for collective exploration, while a shared replay buffer enables collective exploitation. Resource Manager drives this process by dynamically allocating computational resources amongst the learners.}
    \label{fig:cerl_blueprint}
\vspace{-2em}
\end{figure}

In this paper, we introduce Collaborative Evolutionary Reinforcement Learning (CERL), a scalable framework that leverages a portfolio of learners that learn with different time-horizons to explore different parts of the solution space while remaining loyal to the task. This process is directed by a resource manager that dynamically re-distributes computational resources amongst the learners - favoring the best as a form of online algorithm selection. The diverse set of experiences generated by this adaptive process are stored in a shared replay buffer for collective exploration enabling better sample efficiency. 

Figure \ref{fig:cerl_blueprint} illustrates CERL's multi-layered learning approach where each learner exploits the data generated by a diversity of ``behavioral policies" stemming from other learners in the portfolio. An evolutionary population operating in parallel augments this process by extending exploration to the parameter space of policies through mutation. Evolution also introduces redundancies in the population to stabilize learning, intermixes sub-components within policies through crossover, and binds the entire underlying process to generate an emergent learner that exceeds the sum of its parts. Experiments in a range of continuous control benchmarks demonstrate that CERL inherits the best of its composite learners while remaining overall more sample-efficient.

\section{Background}
\label{sec:back}

A standard reinforcement learning setting is often formalized as a Markov Decision Process (MDP) and consists of an agent interacting with an environment over a finite number of discrete time steps. At each time step $t$, the agent observes a state $s_t$ and maps it to an action $a_t$ using its policy $\pi$. The agent receives a scalar reward $r_t$ and transitions to the next state $s_{t+1}$. The process continues until the agent reaches a terminal state marking the end of an episode. The return, $R_t = \sum_{n=1}^{\infty} \gamma^kr_{t+k}$ is the total return from time step $t$ with discount factor $\gamma \in (0, 1]$. The goal of the agent is to maximize this expected return. The state-value function $\mathcal{Q}^\pi(s,a)$ describes the expected return from state $s$ after taking action $a$ and subsequently following policy $\pi$. 

\subsection{Twin Delayed Deep Deterministic Policy Gradients}
\label{sec:back_TD3}
Policy gradient methods re-frame the goal of maximizing the expected return as the minimization of a loss function $L(\theta)$ where $\theta$ encapsulates the agent parameters. A widely used policy gradient method is Deep Deterministic Policy Gradient (DDPG) \cite{lillicrap2015}, a model-free RL algorithm developed for working with continuous, high dimensional actions spaces. Recently, Fujimoto et al. extended DDPG to Twin Delayed DDPG (TD3), \cite{fujimoto2018addressing} addressing the well-known overestimation problem of the former. TD3 was shown to significantly improve upon DDPG and is the state-of-the-art, off-policy algorithm for model-free deep reinforcement learning in continuous action spaces. TD3 uses an actor-critic architecture \cite{sutton1998} maintaining a deterministic policy (actor) $\pi: \mathcal{S} \rightarrow \mathcal{A}$, and two distinct action-value function approximations (critics) $\mathcal{Q}: \mathcal{S} \times \mathcal{A} \rightarrow \mathbb{R}_i$. 

Each critic independently approximates the actor's action-value function $\mathcal{Q}^\pi$. The actor and the critics are parameterized by (deep) neural networks with $\theta^\pi$, $\theta^\mathcal{Q}_a$, and $\theta^\mathcal{Q}_b$ respectively. A separate copy of the actor $\pi'$ and critics: $\mathcal{Q}_a'$ and $\mathcal{Q}_b'$ are kept as target networks for stability. These networks are updated periodically using the actor $\pi$ and critic networks: $\mathcal{Q}_a$ and $\mathcal{Q}_b$ regulated by a weighting parameter $\tau$ and a delayed policy update frequency $d$.   

A behavioral policy is used to explore the environment during training. The behavioral policy is simply a noisy version of the policy: $\pi_b(s) = \pi(s) + \mathcal{N}(0,1)$ where $\mathcal{N}$ is white Gaussian noise. After each action, the tuple ($s_t, a_t, r_t, s_{t+1})$ containing the current state, actor's action, observed reward and the next state, respectively, is saved into a \textbf{replay buffer} $\mathcal{R}$. The actor and critic networks are updated by randomly sampling mini-batches from $\mathcal{R}$. The critic is trained by minimizing the loss function: 

\begin{center}
$L_i = \frac{1}{T} \sum_{i} (y_i - \mathcal{Q}_i(s_i, a_i|\theta^{\mathcal{Q}}))^2$  

where $y_i$ = $r_i$ + $\gamma \displaystyle \min_{j=1,2} \mathcal{Q}_j'(s_{i+1}, _a^\sim|\theta^{\mathcal{Q}_j'})$

where $_a^\sim$ is the noisy action computed by adding Gaussian noise clipped to between $-c$ and $c$. 
$_a^\sim = \pi'(s_{i+1}|\theta^{\pi'}) + \epsilon$,  
$clip \big(\epsilon \sim \mathcal{N}(\mu,\,\sigma^{2})\, -c, c\big)$

\end{center}

This noisy action used for the Bellman update smoothens the value estimate by bootstrapping from similar state-action value estimates. It serves to make the policy smooth and addresses overfitting of the deterministic policy. The actor is trained using the sampled policy gradient:
\begin{center}

$\nabla_{\theta^\pi}J\sim  \frac{1}{T} \sum \nabla_a\mathcal{Q}(s,a|\theta^\mathcal{Q}_a)|_{s=s_i, a=a_i} \nabla_{\theta^\pi} \pi(s|\theta^\pi)|_{s=s_i}$

\end{center}

The sampled policy gradient with respect to the actor's parameters $\theta^\pi$ is computed by backpropagation through the combined actor and critic network. 

\subsection{Evolutionary Algorithms}
Evolutionary algorithms (EAs) are a class of search algorithms characterized by three primary operators: new solution generation, solution alteration and selection \cite{fogel2006,spears1993}. These operations are applied on a population of candidate solutions to continually generate new solutions while retaining promising ones. The selection operation is generally probabilistic, where better solutions with higher fitness values have a higher probability of being selected. Assuming that higher fitness values are representative of good solution quality, the overall quality of solutions will improve with each generation. In this work, each individual in the evolutionary algorithm is a deep neural network representing a policy $\pi$. Mutation is implemented as random perturbations to the weights (genes) of these neural networks. The evolutionary framework used here is closely related to evolving neural networks and is often referred to as neuroevolution \cite{floreano2008,luders2017,risi2017,stanley2002}.

\section{Related Work}

A closely related work to CERL is Population-based Training (PBT) \cite{jaderberg2017}, which employs a population to jointly optimize models and its associated hyperparameters online. However, unlike CERL, PBT does not dynamically redistribute computational resources amongst its learners; instead, it relies entirely on its evolutionary process for learner selection. Additionally, learners in PBT are isolated and do not share experiences with each other for collective exploitation - a key mechanism in CERL for the retention of sample-efficiency. Collective exploitation of a diverse set of experiences is a popular idea, particularly in recent literature. Colas et al. used an evolutionary method (Goal Exploration Process) to generate diverse samples followed by a policy gradient method for fine-tuning the policy parameters \cite{colas2018} while Khadka and Tumer incorporated the two processes to run concurrently formulating a Lamarckian framework \cite{khadka2018evolution}. From an evolutionary perspective, this is closely related to the idea of incorporating learning with evolution \cite{ackley1991,drugan2018,turney1996}. %

Another facet of CERL is algorithm selection \cite{gagliolo2006learning, smith2009cross, rice1976algorithm} - an idea that has been explored extensively in past literature. Lagoudakis and Littman formulated algorithm selection as an MDP and used Q-learning to solve classic order statistic selection and sorting problems \cite{lagoudakis2000algorithm}. Cauwet et al. addressed noisy optimization using a portfolio of online reinforcement learning algorithms \cite{cauwet2014algorithm}. Conversely, Laroche and Feraud introduced Epochal Stochastic
Bandit Algorithm Selection (ESBAS), which tackled algorithm selection in reinforcement learning itself, formulating it as a K-armed stochastic bandit problem \cite{laroche2017reinforcement}. The resource-manager in CERL closely builds on this formulation to inherit its good exploration-exploitation trade-off properties. However, unlike ESBAS, CERL leads to soft algorithm selection - carried out through the allocation of computation resources rather than a hard binary selection.

\section{Motivating Example}
\label{sec:motivating_example}

\begin{figure}

\begin{center}
\includegraphics[width=0.8\columnwidth]{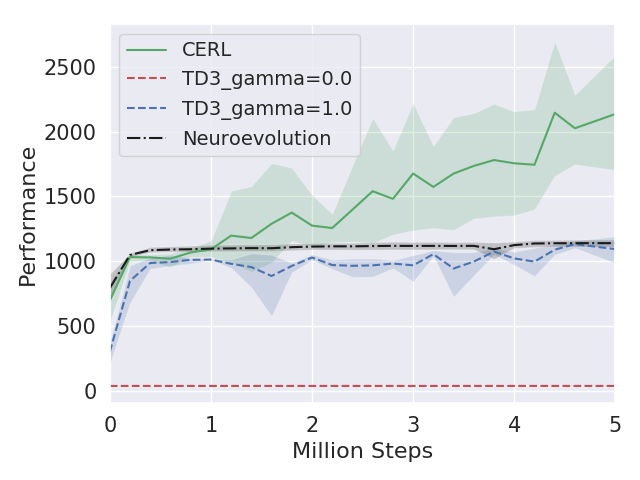}
\caption{Comparative performance of Neuroeovlution, TD3 ($\gamma=0.0, 1.0$) and CERL (built using them) in the Hopper benchmark.}
\label{fig:motivate}
\end{center}
\vspace{-2em}
\end{figure}

Consider the Hopper task from OpenAI gym \cite{brockman2016}, a classic continuous control benchmark used widely in recent DRL literature \cite{duan2016,islam2017,henderson2017,schulman2015b}. Here, the goal is to make a two-dimensional, one-legged robot hop as fast as possible without falling. The task has a state space dimension of $\mathcal{S} = 11$ and action space dimension of $\mathcal{A} = 3$. TD3 has been shown to solve this problem fairly easily \cite{fujimoto2018} (also shown in Figure \ref{fig:mujoco_results} in Section \ref{sec:results}). However, TD3 solves this problem with a tuned discount rate ($\gamma=0.99$). It is interesting how sensitive this performance would be to varying choices of a discount rate ($\gamma$), including ones that are clearly sub-optimal.

 Figure \ref{fig:motivate} shows the comparative performance of 
 TD3 ($\gamma=0.0$), TD3 ($\gamma=1.0$), neuroevolution and our proposed approach: CERL built using the two TD3 variations as its learners.  TD3 ($\gamma=0.0$) represents an extremely greedy learner whose optimization horizon is limited to its immediate reward. On the contrary,  TD3 ($\gamma=1.0$) represents a long-term learner whose optimization horizon is virtually infinite. However, since it seeks to optimize a return which is a function of all future action and states in the trajectory, it learns with significant amount of variance. Both learners represent the extreme ends of the spectrum and would not be expected to learn well. Figure \ref{fig:motivate} corroborates this expectation: TD3 ($\gamma=0.0$) fails to entirely learn as the most greedy action with respect to the immediate reward is rarely aligned with the cumulative episode-wide return. TD3 ($\gamma=1.0$), on the other hand, has a reward ceiling of $1000$ - imposed by the variance of its computed return. Similarly, neuroevolution on its own also fails to solve the task within the 5 million steps tested. However, CERL, which is built directly on top of these learners, is able to continue learning beyond this - reaching a score of $2136\pm512$.
 
While each of the learners fails to solve the problem individually, they collaboratively succeed in solving it under the CERL framework. A key reason here is that each learner fails when required to simultaneously exploit well and produce good behavioral policies that explore the space well. Being able to do both is key to solving the problem and tuning the discount rate is akin to finding this trade-off. CERL provides an alternate approach to finding this trade-off - by employing both learners to explore the space while dynamically distributing the resources to the better performer for effective exploitation. Even when a learner is ill-suited for solving the task by itself, it can serve to be a key 'behavioral policy' that explores critical parts of the search space and generates experiences which are key to learning well on the task. CERL exploits these diversities to define an emergent learner that surpasses the sum of its parts.

\section{Collaborative Evolutionary Reinforcement Learning}
\label{sec:cerl}

\vspace{1em}
\begin{algorithm}[!ht]
\caption{Object Learner}
\label{alg:learner} 
\begin{algorithmic}[1]
\Procedure {Initialize}{$\gamma$}
\State Set discount rate=$\gamma$, $count$=$0$ and value $v$=$0$
\State Initialize actor $\pi$ and critic $\mathcal{Q}$ with weights $\theta^\pi$ and \hspace*{5mm}$\theta^\mathcal{Q}$, respectively    
\State Initialize target actor $\pi'$ and critic $\mathcal{Q}'$ with weights \hspace*{5mm}$\theta^{\pi'}$ and $\theta^{\mathcal{Q}'}$, respectively
\EndProcedure
\end{algorithmic}
\end{algorithm}
\vspace{1em}

\begin{algorithm}[!ht]
\caption{CERL Algorithm}
\label{alg:cerlalgo}
\begin{algorithmic}[1]
\State Initialize portfolio $\mathcal{P}$ with $q$ learners (Alg \ref{alg:learner}) - varying $\gamma$
\State Start allocation $\mathcal{A}$ uniformly, and set \# roll-out $H$ = 0
\State Initialize a population of $k$ actors $pop_{\pi}$ 
\State Initialize an empty cyclic replay buffer $\mathcal{R}$
\State Define a Gaussian noise generator $O =  \mathcal{N}(0, \sigma)$
\State Define a random number generator $r()$ $\in$ $[0,1)$ 

\For{generation = 1, $\infty$}

    \For{actor $\pi$ $\in$ $pop_{\pi}$}
    	\State fitness, R = Evaluate($\pi$, R, noise=None)
    \EndFor
    \State Rank the population based on fitness scores 
    \State Select the first $e$ actors $\pi$ $\in$ $pop_{\pi}$ as elites
    \State Select $(k-e)$ actors $\pi$ from $pop_{\pi}$, to form Set $S$ 
    \hspace*{5mm} using tournament selection with replacement 
    \While{$|S|$ $<$ $(k-e)$}
         \State Use single-point crossover between a randomly \hspace*{5mm}\hspace*{5mm} sampled $\pi \in e$ and $\pi \in S$ and append to $S$
    \EndWhile
    \For{Actor $\pi$ $\in$ Set $S$}
         \If{$r() < mut_{prob}$}
            \State Mutate($\theta^\pi$)
         \EndIf
    \EndFor

    
    \For{Learner $\mathcal{L}$ $\in$ $\mathcal{P}$ }
        \For{ii =1,$\mathcal{A}_i$}
            \State $score,$ R = Evaluate$(L_{\pi}, $R$, noise=O)$
            \State $\mathcal{L}_v$ = $\alpha$ * score + (1 - $\alpha$) * $\mathcal{L}_v$
            \State $\mathcal{L}_{count}$ += 1
        \EndFor
    \EndFor
    
    \State ups = $\#$ of environment steps taken this generation
    \For{ii = 1, ups}
        \For{Learner $\mathcal{L}$ $\in$ $\mathcal{P}$ }
            \State Sample a random minibatch of T transitions \hspace*{5mm}\hspace*{10mm} $(s_i, a_i, r_i, s_{i+1})$ from $\mathcal{R}$
 
            \State Update the critic via a Bellman update \hspace*{5mm}\hspace*{11mm}using the min of $\mathcal{L_Q}_j'(s_{i+1}$ (see \ref{sec:back_TD3})
            
            \State Update $L_\pi$ using the sampled policy \hspace*{5mm}\hspace*{11mm}gradient with noisy actions (see \ref{sec:back_TD3}))
 

            
            
         
             \State Soft update target networks: 
             \State $L_{\theta^{\pi^\prime}} \Leftarrow \tau L_{\theta^\pi} + (1 - \tau) L_{\theta^{\pi^\prime}}$ and  
             \State $L_{\theta^{\mathcal{Q}^\prime}} \Leftarrow \tau L_{\theta^\mathcal{Q}} + (1 - \tau)L_{\theta^{\mathcal{Q}^\prime}}$ 

        \EndFor

    \EndFor
    
    \State Compute the UCB scores $\mathcal{U}$ using 
    \For{Learner $\mathcal{L}$ $\in$ $\mathcal{P}$} 
        \begin{center}
        $\mathcal{U}_i = \mathcal{L}_v + c * \sqrt{\frac{\log_e{H}}{\mathcal{L}_{count}}}$
         \end{center}
    \EndFor
    
    \State Normalize $U$ to be within $[0,1)$ and set $\mathcal{A}$ = []
    \State Sample from $U$ to fill up $\mathcal{A}$

    \If{generation mod $\omega = 0$}
        \For{Learner $\mathcal{L}$ $\in$ $\mathcal{P}$}
            \State Copy $L_\pi$ into the population: for weakest \hspace*{16mm}$\pi \in pop_{\pi}: \theta^\pi \Leftarrow L_{\theta^\pi}$
     \EndFor
    \EndIf
            
\EndFor
\end{algorithmic}
\end{algorithm}

\begin{algorithm}
\caption{Function Evaluate}
\label{alg:episode} 
\begin{algorithmic}[1]
\Procedure {Evaluate}{$\pi$, R, noise}
\State $fitness = 0$
	\State Reset environment and get initial state $s_0$
    \While{env is not done}
        \State Select action $a_t = \pi(s_t|\theta^\pi) + noise_t$ 
        \State Execute action $a_t$ and observe reward $r_t$ and \hspace*{5mm}\hspace*{6mm}new state $s_{t+1}$ 
        \State Append transition $(s_t, a_t, r_t, s_{t+1})$ to $R$ 
        \State $fitness \leftarrow fitness + r_t$ and $s = s_{t+1}$
    \EndWhile
\State Return $fitness$, R
\EndProcedure
\end{algorithmic}
\end{algorithm}

\begin{algorithm}
\caption{Function Mutate}
\label{alg:mutate} 
\begin{algorithmic}[1]
\Procedure {Mutate}{$\theta^\pi$}
\For{Weight Matrix $\mathcal{M} \in \theta^\pi$}
    \For{iteration = 1, $mut_{frac} * |\mathcal{M}|$}
        \State Randomly sample indices $i$ and $j$ from $\mathcal{M}'s$ \hspace*{15mm}first and second axis, respectively
        \If{$r() < supermut_{prob}$} 
            \State $\mathcal{M}[i,j]$ = $\mathcal{M}[i,j]$ * $\mathcal{N}(0,\,100 * \hspace*{20mm}mut_{strength})$
        \ElsIf{$r() < reset_{prob}$}
            \State $\mathcal{M}[i,j]$ = $\mathcal{N}(0,\,1)$
        \Else
            \State $\mathcal{M}[i,j]$ = $\mathcal{M}[i,j]$ * $\mathcal{N}(0,\,mut_{strength})$
        \EndIf
    \EndFor
\EndFor
\EndProcedure
\end{algorithmic}
\end{algorithm}

The principal idea behind Collaborative Evolutionary Reinforcement Learning (CERL) is to incorporate the strengths of multiple learners, each optimizing over varying time-horizons of the underlying task (MDP). While a specific learner is unlikely to be an optimal choice for the task throughout the learning process, a diverse collection of learners is significantly more likely to be so. This is particularly true for exploration, where different learners can contribute a diverse set of behavioral policies while remaining loyal to the task. A shared replay buffer ensures that all learners exploit this diverse data generated. A resource manager supervises this process by dynamically re-distributing computational resources to favor the better performing learners. Finally, this entire underlying apparatus is bound together by evolution which serves to integrate the best policies, explore in the parameter space and exploit any decomposition in the policy space with crossover operands. The emergent learner combines the best of its underlying composite processes, leading to a whole larger than the sum of its parts.

A general flow of the CERL algorithm proceeds as follows: a population of actor networks is initialized with random weights. The population is then \textit{evaluated} in an episode of interaction with the environment (\textit{roll-out}). The fitness for each actor is computed as the cumulative sum of the rewards received in the roll-out. A \textit{selection} operator selects a portion of the population for survival with probability commensurate with their relative fitness scores. The weights of the actors in the population are then probabilistically \textit{perturbed} through mutation and crossover operators to create the next \textbf{generation} of actors. A select portion of actors with the highest relative fitness are shielded from the mutation step and are preserved as elites. 

\textbf{Portfolio:} The procedure described so far is reminiscent of a standard EA. However, in addition to the population of actors, CERL initializes a collection of \textit{learners} (henceforth referred to as a \textit{portfolio}). Each learner is initialized with its own actor, critic and has an associated learning algorithm defined with its own distinct hyperparameters. In this paper, the variation across learners is realized through varying discount rates ($\gamma$). However, in general, this can be any other variation in the hyperparameters, including a difference in the learning algorithm itself. The variation in discount rate used in this work can be interpreted as each learner optimizing over a distinct time-horizon of the underlying MDP. Learners with lower discount rates optimize a ``greedier" objective than the ones with larger discount rates (long-term optimizers). The greedier objective has the benefit of being highly learnable but is not guaranteed to be aligned with the true learning goal. On the other hand, the long-term objective is more aligned to the true learning goal but is not as learnable - suffering from high variance due to its returns being conditioned on a longer time horizon. Thus, the portfolio represents a diverse set of learners, each with its own strengths and weaknesses.

\textbf{Adaptive Resource Allocation:} CERL is initialized with a computation resource budget of $b$ workers dedicated to running roll-outs for its learner portfolio (separate from the resources used to evaluate the evolutionary population of actors). Allocation $\mathcal{A}$ defines the allotment of this resource budget amongst the learners within the portfolio for each generation of learning. This is initialized uniformly - each learner gets an equal number of dedicated workers to run roll-outs using its actor as the behavioral policy. Each learner stores statistics about the number of cumulative roll-outs it has run $y$, and a value metric $v$, defined as the discounted sum of the cumulative returns received from its own roll-outs. $v$ is updated after every roll-out as: 
\begin{center}
       $v' \Leftarrow \alpha * return + (1 - \alpha) * v$ 
\end{center}

After each generation, an upper confidence bound (UCB) \cite{auer2002using} score $\mathcal{U}$ is computed for each learner based on its node statistics using Equation \ref{eqn:ucb}. This formulation is commonly used in solving multi-bandit problems \cite{bubeck2012regret, karnin2013almost}. The UCB score is known to provide good trade-offs between exploitation and exploration and has been extensively used for reinforcement learning in the form of tree searches \cite{anthony2017thinking, silver2016} and algorithms selection \cite{laroche2017reinforcement}. 

\begin{equation}
\label{eqn:ucb}
    \mathcal{U}_i =  v^n_i + c * \sqrt{ \frac{\log( \sum_{i=1}^{b} y_i )} {y_i} }
\end{equation}

\begin{center}
where, $v^n$ is $v$ normalized to be $\in (0,1)$
\end{center}

The UCB scores are normalized to form a probability distribution, and allocation $A$ is re-populated by iterative sampling from this distribution. The allocation describes the new allotment of resources (roll-out workers) amongst the learners for the next generation. The process can be seen as a meta-operation that adaptively distributes resources across the learners dynamically during the course of learning. The underlying UCB technique used to control this distribution ensures a systematic approach to balancing exploitation and exploration when allocating resources across learners.

\textbf{Shared Experiences:} The collective replay buffer is the principal mechanism that enables the sharing of information across the evolutionary population and amongst the learners in the portfolio. In contrast to a standard EA which would extract the fitness metric from each of its roll-outs and disregard them immediately, or ensemble methods that treat different learners separately, CERL pools all experiences defined by the tuple \textit{(current state, action, next state, reward)} in its collective replay buffer. This is done for every interaction, at every time-step, for every episode and for each of its actors (including the evolutionary population and each roll-out conducted by the portfolio of learners). All learners are then able to sample experiences from this collective buffer and use it to update its parameters repeatedly using gradient descent. This mechanism allows for increased information extraction from each individual experiences leading to improved sample efficiency.      

\textbf{Diverse Exploration:} In contrast to most methods where a learner learns based on data that its behavioral policy generates, CERL enables its portfolio of learners to leverage the data generated by a diverse set of actors. This includes the actors within the neuroevolutionary population and the actors stemming from other learners in the portfolio. Since each learner optimizes over varying time-horizons of the same underlying MDP, the associated actors lead to diverse behavioral policies exploring different regions of the solution space while remaining aligned with the task at hand. Additionally, in contrast to the learners which explore in their \textit{action} space, the neuroevolutionary population explores in its \textit{parameter} (neural weights) space using the mutation operator. The two processes complement each other and collectively lead to an effective strategy that is able to better explore the policy space.

\textbf{Portfolio $\rightarrow$ EA:} Periodically, each learner network is copied into the evolutionary population of actors, a process referred to as \textit{Lamarckian transfer}. The frequency of \textit{Lamarckian transfer} controls the flow of information from the gradient-based learners in the portfolio to the gradient-free evolutionary population. This is the core mechanism that enables the evolutionary framework to directly leverage the information learned through gradient-based optimization. The evolutionary process also acts as an amplifier in the realization of adaptive resource allocation. Good learner policies are selected to survive and reproduce - extending their influence in the population over subsequent generations. These policies and their descendants contribute increasingly more data experiences into the collective replay buffer and influence the learning of the all portfolio learners. Bad learner policies, on the other hand, are rejected to minimize their influence. Finally, crossover serves to exploit any decomposability in the policy space and combines good ``sub-components of the policies" present in the diverse evolutionary population.

Algorithm \ref{alg:cerlalgo}, \ref{alg:episode} and \ref{alg:mutate} provide a detailed pseudo-code of the CERL algorithm using a portfolio of TD3 learners. The choice of hyperparameters is explained in the Appendix. Additionally, our source code \footnote{\href{https://github.com/intelai/cerl}{github.com/intelai/cerl}} is available online.

\section{Results}
\label{sec:results}

\textbf{Domain:} CERL is evaluated on $5$ continuous control tasks on Mujoco \cite{todorov2012}. These benchmarks are used widely in the field \cite{khadka2018evolution,such2017,schulman2017} and are hosted on OpenAI gym \cite{brockman2016}.    

\textbf{Compared Baselines:} For each benchmark, we compare the performance of CERL with its composite learners ran in isolation. While not constrained to this arrangement, CERL here is built using a combination of a neuroevolutionary algorithm (EA) and $4$ policy gradient based learners. We use TD3 \cite{fujimoto2018addressing} as our policy gradient learner as it is the current state-of-the-art off-policy algorithm for these benchmarks. The $4$ TD3 learners are identical with each other apart from their discount rates which are $0.9$, $0.99$, $0.997$, and $0.9995$. These were not tuned for performance.

We also ran CERL with a single learner - picking the best TD3 learner for each task. This is equivalent to ERL \cite{khadka2018evolution} with the exception of the resource manager. However, the resource manager does not have any functional effect when there is only one learner.

\textbf{Methodology for Reported Metrics:} 
For TD3, the actor network was periodically tested on $10$ task instances without any exploratory noise. The average score was logged as its performance. During each training generation, the actor network with the highest fitness was selected as the champion. The champion was then tested on $10$ task instances, and the average score was logged. This protocol shielded the reported metrics from any bias of the population size. We conduct $5$ statistically independent runs with random seeds from $\{2018, 2022\}$ and report the average with error bars showing a $95\%$ confidence interval.

\textbf{The ``Steps" Metric:} All scores reported are compared against the number of environment steps. A step is defined as an agent taking an action and receiving a reward back from the environment. To make the comparisons fair across single-agent and population-based algorithms, all steps taken by all actors in the population, and by all learners in the portfolio are counted cumulatively.

\textbf{Hyperparameter Selection:} The hyperparameters used for CERL were \textbf{not} tuned to generate the results, unless specifically stated. The parameters used for the TD3 learners were simply inherited from \cite{fujimoto2018addressing}, while the evolutionary parameters were inherited from \cite{khadka2018evolution}. The computational budget of \textit{b} workers was set to $10$ to match the evolutionary population size. The UCB exploration coefficient was set to $0.9$ which numerically makes the relative weight of exploration and exploitation terms in Equation \ref{eqn:ucb} close to equilibrium at the start.  

\begin{figure}[t]

\includegraphics[width=0.9\columnwidth]{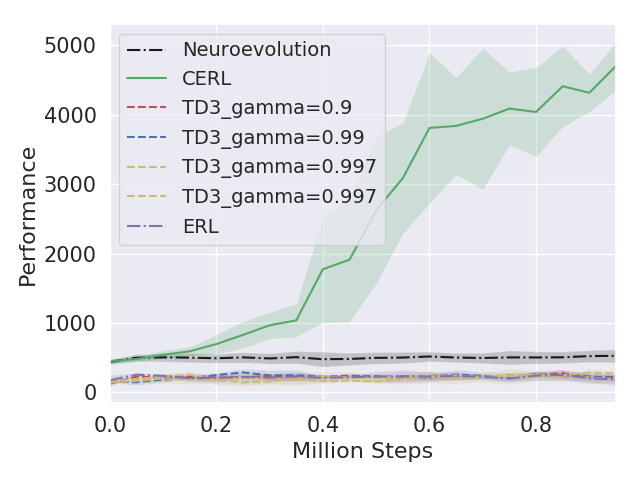}
\caption{Comparative Results for CERL tested against its composite learners in the Humanoid benchmark.}
\label{fig:humanoid}
\vspace{-1em}
\end{figure}

\textbf{Humanoid:} Figure \ref{fig:humanoid} shows the comparative performance of CERL, alongside its composite learners. CERL significantly outperforms neuroevolution, as well as all versions of TD3 with varying discount rates. The TD3 learners fail to learn at all, which is consistent with reports in previous literature \cite{haarnoja2018}. On the other hand, neuroevolution alone was shown to solve Humanoid, but required 62.5 millions roll-outs \cite{lehman2018more}. CERL is able to achieve a score of $4702.0\pm356.5$ within $1$ million environment steps (approximately ~4000 roll-outs). Considering that CERL only uses a combination of these learners, this is a significant result. Each learner in isolation fails to learn on the task entirely, while the same learners when incorporated under the CERL framework, are able to solve it jointly. This is because none of the learners are able to succeed when burdened with both exploring the solution space to generate an expansive set of data, and exploiting it aptly. However, when the learners collectively explore diverse regions of the solution space, and collectively exploit these experiences, they succeed. The single-learner ERL also fails to learn this task. Since the key difference between ERL and CERL is the use of multiple learners, this demonstrates that the performance gains of CERL come primarily from this collaboration.

\begin{figure}[t]
\includegraphics[width=0.9\columnwidth]{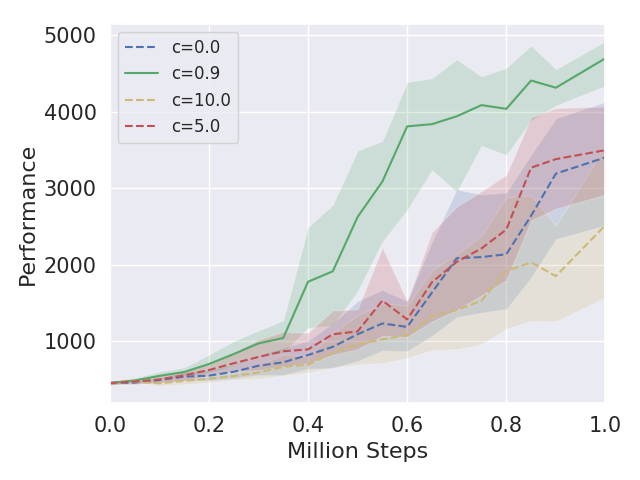}
\caption{Sensitivity analysis for resource-manager exploration (c) in the Humanoid benchmark}
\label{fig:hum_c}
\vspace{-1em}
\end{figure}

\textbf{Resource-manager's Sensitivity to Exploration}: Figure \ref{fig:hum_c} shows the comparative performance for CERL tested with varying $c$ (exploration coefficient in Equation \ref{eqn:ucb}) for the Humanoid benchmark. CERL with $c=0.9$ performs the best as it provides a good balance of exploration and exploitation for the resource-manager. However, CERL with $c=0.0$ and $c=5.0$ both are also able to learn well the benchmark, but are less sample-efficient. An important point to note is that  $c=0.0$ does not lead to the complete lack of exploration. As all learners start with random weights, the returns are close to random at the beginning of learning and serves to bootstrap exploration. On the other hand, a $c$ of $10$ does lead to extremely high exploration. As expected, this prolonged exploration leads to even lower sample efficiency. This highlights the role that the resource-manager plays in dynamically redistributing resource and finding the balance between exploration and exploitation.   

\begin{figure*}[ht!]
\subfigure[Hopper]{
\includegraphics[width=0.49\columnwidth]{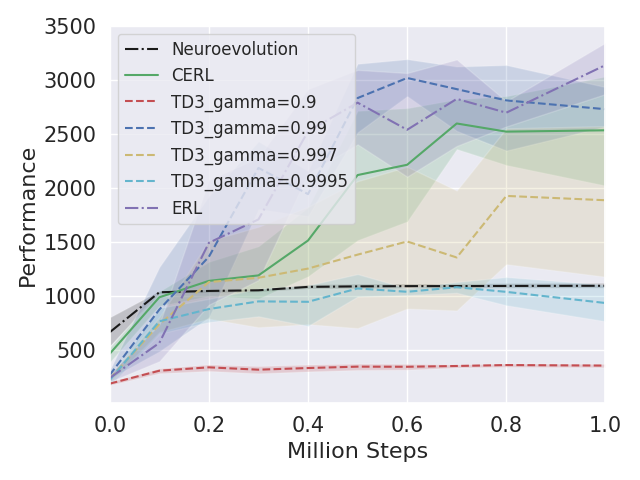}}\hspace*{\fill}
\subfigure[Swimmer]{
\includegraphics[width=0.49\columnwidth]{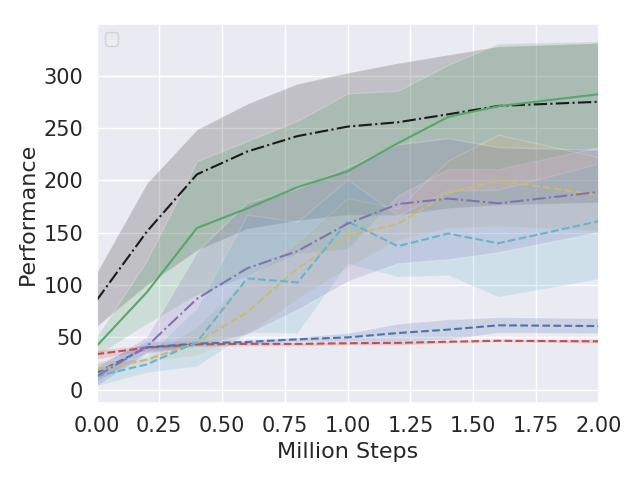}}\hspace*{\fill}
\subfigure[HalfCheetah]{
\includegraphics[width=0.49\columnwidth]{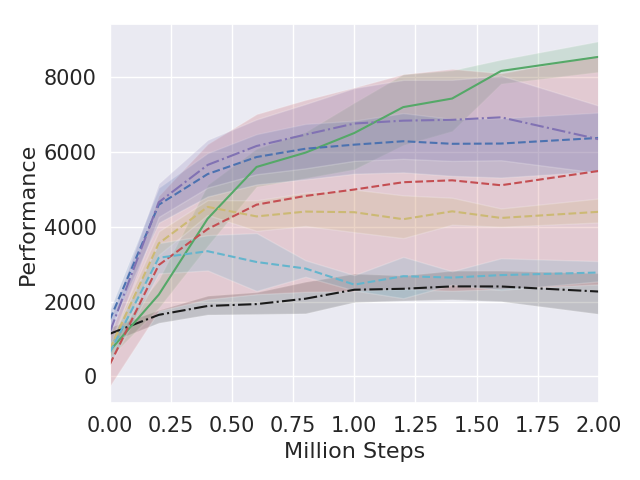}}
\subfigure[Walker2D]{
\includegraphics[width=0.49\columnwidth]{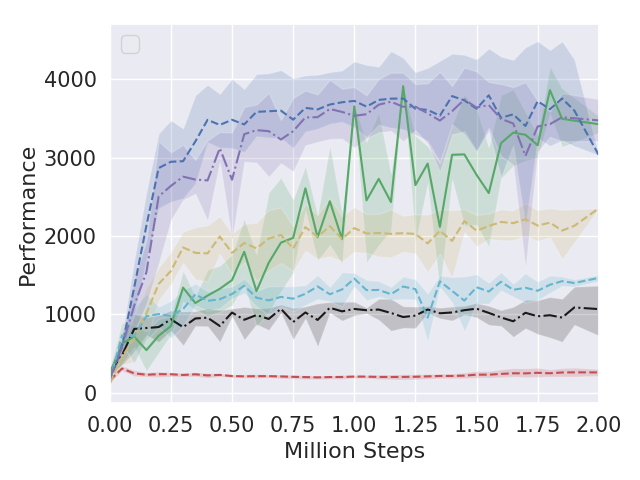}}\hspace*{\fill}
\caption{Comparative results of CERL with 4 learners (TD3 with discount rates of 0.9, 0.99, 0.997 and 0.9995) against the learners in isolation, and neuroevolution.}
\label{fig:mujoco_results}
\vspace{-1em}
\end{figure*}

\textbf{Additional Mujoco Experiments:} Figure \ref{fig:mujoco_results} shows the comparative performance of CERL, alongside its composite learners in $4$ additional environments simulated using Mujoco. Unlike the 3D humanoid benchmark, these domains are 2D, have considerably smaller state and action spaces, and are relatively simpler to solve. One of the four TD3 learners: TD3 with a discount rate of 0.99 (TD3-0.99) is able to solve 3 out of the 4 benchmarks, with the exception of Swimmer. CERL is also able to solve these benchmarks but is less sample-efficient that TD3-0.99. However, on the Swimmer benchmark, while all of the TD3 learners fail to solve the task, CERL successfully solves it similar to neuroevolution. This emphasizes the key strength of CERL: the ability to inherit the best of its composite approaches. 

While TD3-0.99 is more sample-efficient in $3$ out of the $4$ benchmarks, CERL is more sample-efficient than all the other TD3-based learners. This suggests that $0.99$ is an ideal discount rate for these tasks. Any deviation from this value leads to considerable loss in performance for TD3. In other words, this is a sensitive hyperparameter that has to be rigorously tuned. CERL achieves this functionality online through its resource-manager, which adaptively re-distributes computational resources across the learners. While this invariably leads to the use of more samples when compared to an ideal hyperparameter that is known \textit{a priori}, CERL is able to identify and exploit the best hyperparameters online via joint exploration. Additionally, as demonstrated in the cases of Swimmer and Humanoid (Figure \ref{fig:humanoid}), this exploration itself is critical to successful learning as there does not exist one hyperparameter that can solve the task all by itself. Overall, CERL enables an arguably simpler alternative to network design compared to complex hyperparameter tuning methodologies. 

\textbf{Allocation:} Table \ref{alloc_table} reports the final cumulative resource-allocation rate across the four learners for CERL in the five Mujoco benchmarks tested. L1, L2, L3 and L4 correspond to learners with $\gamma=0.9, 0.99, 0.997$ and $0.9995$, respectively. L2 seems to be the learner that is generally preferred across most tasks. This is not surprising as this value for $\gamma$ is the hyperparameter used in \cite{fujimoto2018addressing} after tuning. However, in the Swimmer benchmark, this choice of hyperparameter is not ideal. Learners with higher $\gamma$ perform significantly better on the task (Figure \ref{fig:mujoco_results}). CERL is able to identify this online and allocates more resources to L3 and L4 with higher $\gamma$. This flexibility for online algorithm selection, in combination with its evolutionary population, enables CERL to solve the Swimmer benchmark effectively.

\begin{table}
\centering
\caption{Average cumulative resource-allocation rate for CERL across benchmarks. (error intervals omitted as all were $<$ 0.04)}
\vspace{1em}
\setlength\tabcolsep{10.0pt}
 \begin{tabular}{|c|| c| c|c|c|} 
  \hline
Task & L1 & L2 & L3 & L4 \\ [0.5ex] 
 \hline\hline
 Humanoid & $0.24$ & $0.35$ & $0.20$ & $0.20$ \\ [0.5ex] 
 \hline
 Hopper & $0.14$ & $0.27$ & $0.32$ & $0.27$ \\ [0.5ex] 
 \hline
 Swimmer& $0.17$ & $0.20$ & $0.36$ & $0.27$ \\ [0.5ex] 
 \hline
 HalfCheetah & $0.29$ & $0.32$ & $0.24$ & $0.15$ \\ [0.5ex] 
 \hline
Walker & $0.14$ & $0.28$ & $0.33$ & $0.25$ \\ [0.5ex] 
 \hline

 \hline
\end{tabular}
\label{alloc_table}
\vspace{-1em}
\end{table}

\section{Discussion}
\label{sec:discussion}
We presented CERL, a scalable platform that allows gradient-based learners to jointly explore and exploit solutions in a gradient-free evolutionary framework. Experiments in continuous control demonstrate that CERL's emergent learner can outperform its composite learners while remaining overall sample-efficient compared to traditional approaches.

\textbf{Strengths}: CERL is generally insensitive to its hyperparameters and to those of the individual learners. The Humanoid and Swimmer problems are examples where state-of-the-art algorithms show high sensitivity to their hyperparameters while CERL required no hyperparameter tuning. Significantly, the Humanoid problem demonstrates that CERL is able to find effective solutions using participating learners that fail completely on their own. This makes CERL a simpler design alternative to complex hyperparameter tuning and one that seems to generalize well across multiple tasks. 

A practical consideration for CERL is the parallel operation of gradient-based and gradient-free methods. The former, involving backpropagation, is typically suited for GPUs. The latter, involving forward-propagation, is typically suited for CPUs and is highly scalable, leading to impressive wall-clock performances \cite{salimans2017,such2017}. By leveraging both modes simultaneously, CERL provides a principled way to parallelize learning and to cater one's learning algorithm to the available hardware.

\textbf{Limitations}: CERL can be less sample-efficient for simple tasks where the ideal hyperparameters are known \textit{a priori}. This is apparent in the case of Walker2d (Fig \ref{fig:mujoco_results}) and can be attributed to the exploration involved in selecting learners. However, CERL does eventually match the performance shown by the learner with the known ideal hyperparameter. This weakness of CERL is contingent on the ability to derive the ideal parameters for a learner - a process which by itself generally consumes significant resources that are often not reported in literature. 

\textbf{Future Work}: Here, we explored homogeneous learners optimizing over varying time-horizons of a task. Future work will extend this to learners that are different algorithms themselves. Incorporating stochastic actors from SAC \cite{haarnoja2018} with the deterministic TD3 actors is an exciting area. Another promising line of work would be to incorporate learning within the resource manager to augment the current UCB formulation. 
\bibliography{cerl}
\bibliographystyle{icml2019}
\pagebreak

\section*{Appendix}


\begin{table}[h]
\caption{Hyperparameters for CERL}
\begin{tabular}{|c||c|} 
 \hline
 Hyperparameter & Value\\ 
 \hline \hline
  Population size $k$ & 10 \\ 
   Roll-out size $b$ & 10 \\ 
    Target weight $\tau$  & $5e^{-3}$ \\ 
      Actor Learning Rate   & $1e^{-3}$ \\ 
      Critic Learning Rate   & $1e^{-3}$ \\ 
         Replay Buffer   &  $1e^{6}$\\ 
           Batch Size    & $256$ \\ 
              Mutation Probability $mut_{prob}$    & $0.9$ \\ 
    Mutation Fraction $mut_{frac}$   &  $0.1$\\ 
   Mutation Strength $mut_{strength}$  & $0.1$ \\          
            Super Mutation Probability $supermut_{prob}$        & $0.05$ \\ 
  Reset Mutation Probability $resetmut_{prob}$ & $0.05$ \\ 
 Number of elites $e$ & $k/5$ \\ 
Lamarckian Transfer Period $\omega$  & $5$ \\ 
Value Learning $\alpha$  & $0.2$ \\ 
 Normalized Observation &  None \\       
               Gradient Clipping   & None \\ 
        Exploration Policy  & $\mathcal{N}(0, \sigma)$ \\ 
        Exploration Noise $\sigma$  & $0.1$ \\ 

 \hline \hline
\end{tabular}
\label{varied}
\vspace{2em}
\end{table}

This section details the hyperparameters used for Collaborative Evolutionary Reinforcement Learning (CERL) across  all benchmarks. 

\begin{itemize}

\item \textbf{Optimizer = Adam} \\
Adam optimizer was used to update both the actor and critic networks for all learners.

\item \textbf{Population size $k$ =  10} \\This parameter controls the number of different individual actors (policies) that are present in the evolutionary population. 

\item \textbf{Roll-out size $b$ =  10} \\This parameter controls the number of roll-out workers (each running an episode of the task) that compose the computational resource available to the resource-manager. 

\textbf{Note:} The two parameters above (population size $k$ and roll-out size $b$) collectively modulates the proportion of exploration carried out through noise in the actor's \textit{parameter} space and its \textit{action} space.

\item \textbf{Target weight $\tau = 5e^{-3}$} \\This parameter controls the magnitude of the soft update between the actors and critic networks, and their target counterparts.

\item \textbf{Actor Learning Rate = $1e^{-3}$} \\
This parameter controls the learning rate of the actor network.

\item \textbf{Critic Learning Rate = $1e^{-3}$} \\
This parameter controls the learning rate of the critic network.

\item \textbf{Replay Buffer Size = $1e^6$} \\
This parameter controls the size of the replay buffer. After the buffer is filled, the oldest experiences are deleted in order to make room for new ones.

\item \textbf{Batch Size = $256$} \\
This parameters controls the batch size used to compute the gradients.

\item \textbf{Actor Neural Architecture = $[400, 300]$} \\
The actor network consists of two hidden layers, each with $400$ and $300$ nodes, respectively. Exponential Linear Units (ELU) was used as the activation function. Layer normalization was used before each layer.

\item \textbf{Critic Neural Architecture = $[400, 300]$} \\
The actor network consists of two hidden layers, each with $400$ and $300$ nodes, respectively. Exponential Linear Units (ELU) was used as the activation function. Layer normalization was used before each layer.

\item \textbf{Number of Elites $e$ = $k/5$} \\
The number of elites was set to be $20\%$ of the population size $(k)$. This parameter controls the fraction of the population that are categorized as elites. Since an elite individual (actor) is shielded from the mutation step and preserved as it is, the elite fraction modulates the degree of exploration/exploitation within the evolutionary population. In general, tasks with more stochastic dynamics (correlating with more contact points) have a higher variance in fitness values. A higher elite fraction in these tasks helps in reducing the probability of losing good actors due to high variance in fitness, promoting stable learning.  

\item \textbf{Mutation Probability $mut_{prob} = 0.9$} \\
This parameter represents the probability that an actor goes through a mutation operation between generation.

\item \textbf{Mutation Fraction $mut_{frac} = 0.1$} \\
This parameter controls the fraction of the weights in a chosen actor (neural network) that are mutated, once the actor is chosen for mutation.

\item \textbf{Mutation Strength $mut_{strength} = 0.1$} \\
This parameter controls the standard deviation of the Gaussian operation that comprises mutation.

\item \textbf{Super Mutation Probability $supermut_{prob} = 0.05$} \\
This parameter controls the probability that a super mutation (larger mutation) happens in place of a standard mutation.

\item \textbf{Reset Mutation Probability $resetmut_{prob} = 0.05$} \\
This parameter controls the probability a neural weight is instead reset between $\mathcal{N}(0,1)$ rather than being mutated.

\item \textbf{Value learning rate $\alpha = 0.2$ } \\
This parameter controls the learning rate used to update the value of a learner after receiving a fitness value for its roll-out. 

\item \textbf{Lamarckian Transfer Period $\omega = 5$} \\
This parameter controls the frequency of information flow between the portfolio of learners and the evolutionary population. A higher $\omega$ generally allows more time for expansive exploration by the evolutionary population while a lower $\omega$ can allow for a more narrower search. The parameter controls how frequently the exploration in action space (portfolio of gradient-based learners) shares information with the exploration in the parameter space (actors in the evolutionary population).

\item \textbf{Exploration Noise $\sigma = 0.1$} \\
This parameter controls the standard deviation of the Gaussian operation that comprise the noise added to the actor's actions during exploration by the learners (learner roll-outs). 

\end{itemize}

\end{document}